\documentclass{article}

\usepackage{arxiv}

\usepackage[utf8]{inputenc} 
\usepackage[T1]{fontenc}    
\usepackage{hyperref}       
\usepackage{url}            
\usepackage{booktabs}       
\usepackage{amsfonts}       
\usepackage{nicefrac}       
\usepackage{microtype}      
\usepackage{lipsum}
\usepackage{graphicx}
\graphicspath{ {./images/} }
\usepackage{bm}
 \usepackage{amsmath}
 \usepackage{overpic}
 \usepackage{array}
 \usepackage{multirow}
 \usepackage{booktabs}
\usepackage{tabularx}
\usepackage{tabularray}
\usepackage[table]{xcolor}
\usepackage{color,colortbl}
\definecolor{LightBlue}{rgb}{0.88,0.9,0.9}
\colorlet{LightBlue}{LightBlue!40}

\title{Scanner-Agnostic MRI Harmonization via SSIM-Guided Disentanglement}

\author{
 Luca Caldera \\
  MOX, Department of Mathematics \\
  Politecnico di Milano \\
  Milan, Italy \\
  \texttt{luca.caldera@polimi.it} \\
   \And
 Lara Cavinato \\
  MOX, Department of Mathematics \\
  Politecnico di Milano \\
  Milan, Italy \\
  \texttt{lara.cavinato@polimi.it} \\
  \And 
 Francesca Ieva \\
  MOX, Department of Mathematics \\
  Politecnico di Milano \\
  Milan, Italy \\
  \texttt{francesca.ieva@polimi.it} \\
  \And 
 For the Alzheimer’s Disease Neuroimaging Initiative\thanks{%
Data used in preparation of this article were obtained from the Alzheimer’s
Disease Neuroimaging Initiative (ADNI) database (\url{adni.loni.usc.edu}). 
As such, the investigators within ADNI contributed to the design and implementation 
of ADNI and/or provided data but did not participate in analysis or writing of this report. 
A complete listing of ADNI investigators can be found at: 
\url{http://adni.loni.usc.edu/wp-content/uploads/how_to_apply/ADNI_Acknowledgement_List.pdf}.
} \\
}

\begin{document}
\maketitle
\begin{abstract}
The variability introduced by differences in MRI scanner models, acquisition protocols, and imaging sites hinders consistent analysis and generalizability across multicenter studies. We present a novel image-based harmonization framework for 3D T1-weighted brain MRI, which disentangles anatomical content from scanner- and site-specific variations. The model incorporates a differentiable loss based on the Structural Similarity Index (SSIM) to preserve biologically meaningful features while reducing inter-site variability. This loss enables separate evaluation of image luminance, contrast, and structural components. Training and validation were performed on multiple publicly available datasets spanning diverse scanners and sites, with testing on both healthy and clinical populations. Harmonization using multiple style targets, including style-agnostic references, produced consistent and high-quality outputs. Visual comparisons, voxel intensity distributions, and SSIM-based metrics demonstrated that harmonized images achieved strong alignment across acquisition settings while maintaining anatomical fidelity. Following harmonization, structural SSIM reached 0.97, luminance SSIM ranged from 0.98 to 0.99, and Wasserstein distances between mean voxel intensity distributions decreased substantially. Downstream tasks showed substantial improvements: mean absolute error for brain age prediction decreased from 5.36 to 3.30 years, and Alzheimer’s disease classification AUC increased from 0.78 to 0.85. Overall, our framework enhances cross-site image consistency, preserves anatomical fidelity, and improves downstream model performance, providing a robust and generalizable solution for large-scale multicenter neuroimaging studies.
\end{abstract}



\section{Introduction}
\label{sec:SCIENTIFIC-BACKGROUND}
The growing availability of brain MRI datasets presents valuable opportunities for understanding neurological diseases and supporting clinical applications. However, differences in imaging protocols, scanner models, and acquisition settings introduce inconsistencies that compromise the reliability of imaging biomarkers. Scanner type and acquisition site most strongly influence image characteristics such as contrast, brightness, and spatial resolution. Even identical scanner models can yield varying results due to differences in hardware, software, and maintenance, complicating multicenter studies and reducing reproducibility \cite{takao2011effect,shinohara2017volumetric}. Therefore, harmonizing MRI data across scanners and sites is essential for consistent, comparable analyses.

Harmonization techniques fall into two categories: feature-based and image-based. Feature-based methods adjust derived metrics—such as cortical thickness, regional volumes, or diffusion metrics—without altering raw images \cite{torbati2021multi,radua2020increased,moyer2020scanner}. 
Although computationally efficient, these methods depend heavily on preprocessing pipelines and risk propagating site-specific biases, which limit their ability to generalize across scanners and capture fine-grained voxel-level variations \cite{abbasi2024deep}.

To overcome these limitations, image-based methods leverage machine learning to directly correct scanner- and site-induced variability at the voxel level. These approaches include Transformer models, Image-to-Image (I2I) translation, and Style Transfer approaches. Deep learning methods like CALAMITI \cite{zuo2021unsupervised}, MURD \cite{liu2024learning}, IGUANe \cite{roca2025iguane}, STGAN \cite{choi2020stargan}, and DISARM \cite{caldera2025disarm, caldera2025disarm++} have shown promise in reducing site-related variability and improving harmonization for multicenter MRI studies.
Although image-based methods directly address voxel-level variability, they often struggle to balance structural fidelity and visual consistency. Some approaches overemphasize appearance similarity, potentially altering anatomical content, while others preserve structure too rigidly, limiting the extent of harmonization. In addition, many models operate on 2D slices or small patches, which constrains their ability to capture global 3D context.

In this study, we propose a novel 3D MRI harmonization framework that extends the AGUIT domain translation network \cite{li2019attribute}—originally designed for 2D natural images—to volumetric medical data. This adaptation enables the disentanglement of anatomical structure from scanner- and site-specific variations, allowing the generation of harmonized images that preserve biologically meaningful anatomy while substantially reducing variability across imaging sites.
The main contributions of this work are threefold: (1) we design a new harmonization model tailored for medical imaging; (2) 
we propose a differentiable loss function based on the Structural Similarity Index (SSIM) \cite{wang2004image}, allowing for end-to-end optimization that enhances perceptual consistency, preserves anatomical fidelity, and promotes appearance consistency across harmonized images; and (3) we conduct extensive validation on multiple publicly available datasets encompassing diverse scanners, sites, and populations. Validation and testing included both healthy individuals and clinical cohorts, such as autism spectrum disorder (ASD) and Alzheimer’s disease (AD) groups. In total, our evaluation comprised 2,733 3D T1-weighted MRI volumes acquired from 88 imaging sites using 19 different scanner models from the three main manufacturers—Siemens, Philips, and GE—providing highly heterogeneous and representative testing conditions that demonstrate the robustness and generalizability of the proposed framework.

The remainder of this paper is organized as follows: Section 2 describes the proposed harmonization framework; Section 3 outlines the datasets and image preprocessing steps; Section 4 details the experimental setup; Section 5 reports and discusses the experimental results; and Section 6 concludes the paper and outlines future directions.

\section{Methodology}
\label{sec:METH}
This section presents the proposed harmonization framework. We first introduce the mathematical formulation of the problem, followed by a detailed description of the model architecture and the training procedure.

\subsection{Mathematical Formulation}
\label{subsec:MATH}
Let $\mathcal{X}_u$, $\mathcal{X}_l \in \mathbb{R}^{1 \times H \times W \times D}$ denote the sets of unlabeled and labeled MR images, respectively. Each labeled image in $\mathcal{X}_l$ is associated with a label vector $\bm{l} \in \mathcal{L} = \{-1, 1\}^K$, where $K$ is the number of scanners and sites considered. In each label vector, the entry corresponding to the scanner and site where the image was acquired is set to 1, with all other entries set to -1. 
We assume that each image can be disentangled into two latent spaces:
\begin{itemize}
    \item the anatomical space $\mathcal{B}$, which encodes the structural information of the brain, and
    \item the style space $\mathcal{S}$, which captures the non-structural features of the MR scan.
\end{itemize}
Formally, an image $\bm{x} \in \mathcal{X}$ can be represented by a pair of latent codes, where $\bm{c} \in \mathcal{B}$ and $\bm{s} \in \mathcal{S}$. The style code $\bm{s}$ may itself be decomposed into a noise component $\bm{n} \in \mathcal{N}$ and a label component $\bm{l} \in \mathcal{L}$, such that $\bm{s} = (\bm{n}, \bm{l})$, where the label component $\bm{l}$ captures scanner- and site-specific variations, while the noise component $\bm{n}$ accounts for subject-specific, non-anatomical variations that introduce residual inconsistencies in MRI appearance possibly beyond scanner- or site-related effects.

\subsection{Model Architecture}
\label{subsec:ARCHITECTURE}
The network architecture comprises two encoders and a generator. The brain structure encoder, \( E_b: \mathcal{X} \rightarrow \mathcal{B} \), maps an input image \( \bm{x} \in \mathcal{X} \) to a lower-dimensional space \( \mathcal{B} \), capturing anatomical features. The style encoder, \( E_s: \mathcal{X} \rightarrow \mathcal{S} \), extracts non-anatomical features from the same image \( \bm{x} \), encoding them into a separate latent space \( \mathcal{S} \). The complete style vector \( \bm{s} \in \mathcal{S} = (\mathcal{N}, \mathcal{L}) \) is constructed by combining the noise component with the label component \( \bm{l} \in \mathcal{L} \), which contains information about the scanner and the site of image acquisition. Finally, the generator, \( G: (\mathcal{B}, \mathcal{S}) \rightarrow \mathcal{X} \), synthesizes a new image \( \hat{\bm{x}} \in \mathcal{X} \) that retains the anatomical structure from \( \mathcal{B} \) while incorporating the scanner-site style attributes from \( \mathcal{S} \). Additionally, adversarial training is supported by incorporating discriminators to enhance realism and scanner-site-specific characteristics alignment.

\subsection{Training}
\label{subsec:TRAINING}

\begin{figure}[t!]
    \centering
    \includegraphics[width=\linewidth]{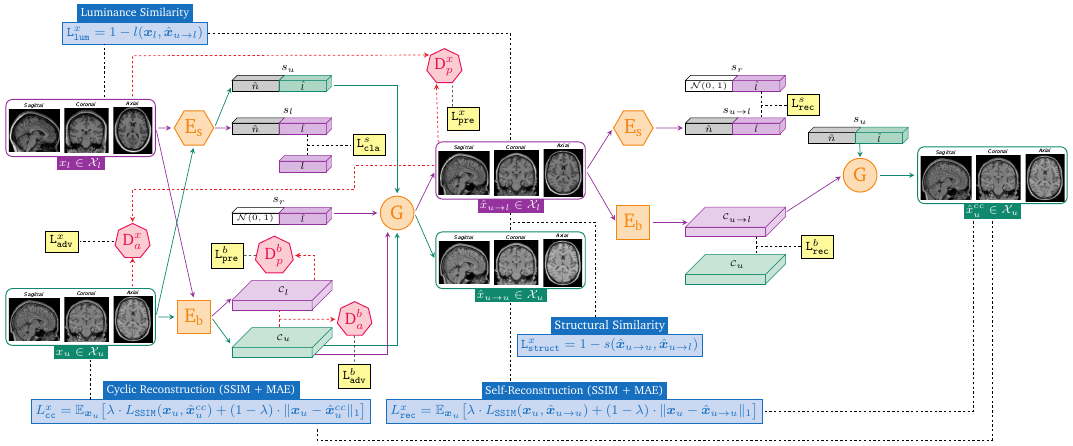}
    \caption{A high-level functional diagram of the model training procedure.}
    \label{train_proc}
\end{figure}

The training process, illustrated in Figure \ref{train_proc}, is performed on pairs of unlabeled and labeled images, denoted as $\bm{x}_u \in \mathcal{X}_u$ and $\bm{x}_l \in \mathcal{X}_l$, respectively. Each image is independently processed by the encoders, $E_b$ and $E_s$, to extract latent representations: the style codes $\bm{s}_u$ and $\bm{s}_l$, and the anatomical structure codes $\bm{c}_u$ and $\bm{c}_l$. The image $\bm{x}_u$ is reconstructed by passing its brain structure code $\bm{c}_u$ and style code $\bm{s}_u$ to the generator. To translate scanner-site-specific characteristics, the generator is instead given $\bm{c}_u$ and a style code $\bm{s}_r$, where $\bm{s}_r$ is composed of random noise and the scanner-site label $l$ from $\bm{x}_l$. The resulting generated images are denoted as $\hat{\bm{x}}_{u \rightarrow u}$ and $\hat{\bm{x}}_{u \rightarrow l}$, respectively. For cyclic reconstruction, the encoders reprocess $\hat{\bm{x}}_{u \rightarrow l}$ to extract its style and brain structure codes. The brain structure code $\bm{c}_{u \rightarrow l}$ is then passed into the generator along with the style vector $\bm{s}_u$ of $\bm{x}_u$. We denote the reconstructed image as $\hat{\bm{x}}_{u}^{cc}$. The training objective involves minimizing a combination of loss functions, each guiding specific subnetworks in the architecture, as outlined in Figure \ref{train_proc}. 

We devise a new Structural Similarity Index Measure (SSIM)-based loss formulation that allows separate evaluation of the image luminance \(l(x, y)\), contrast \(c(x, y)\), and structure \(s(x, y)\) components:
\[
L_{\text{SSIM}}(x, y) = 1 - \text{SSIM}(x, y) = 1 - l(x, y)^\alpha \cdot c(x, y)^\beta \cdot s(x, y)^\gamma \notag
\]
with
\begin{align}
l(x, y) &= \frac{2\mu_x \mu_y + C_1}{\mu_x^2 + \mu_y^2 + C_1}, \quad 
c(x, y) = \frac{2\sigma_x \sigma_y + C_2}{\sigma_x^2 + \sigma_y^2 + C_2}, \quad 
s(x, y) = \frac{\sigma_{xy} + C_3}{\sigma_x \sigma_y + C_3} \notag
\end{align}

where \( \mu_x, \mu_y \) denote image means; \( \sigma_x, \sigma_y \), their standard deviations; and \( \sigma_{xy} \), their covariance. Constants \( C_1, C_2, C_3 \) stabilize computation, while exponents \( \alpha, \beta, \gamma \) weight the SSIM components, allowing tailored loss functions that emphasize luminance, contrast, or structure.

We thus further combine the SSIM with Mean absolute error (MAE) in the Cycle Consistency Loss $L_{\text{cc}}^{x}$ to ensure consistency in cyclic image translation, and in the Reconstruction Loss $L_{\text{rec}}^{x}$ to ensure faithful reconstruction of the input image, with \( \lambda \) being a weighting factor:

\begin{align*}
L_{\text{cc}}^{x} & = \mathbb{E}_{\bm{x}_u} \big[ 
\lambda \cdot L_{\text{SSIM}}(\bm{x}_u, \hat{\bm{x}}_{u}^{cc})
+ (1 - \lambda) \cdot \lVert \bm{x}_u - \hat{\bm{x}}_{u}^{cc} \rVert_1 
\big] \\
L_{\text{rec}}^{x} & = \mathbb{E}_{\bm{x}_u} \big[ 
\lambda \cdot L_{\text{SSIM}}(\bm{x}_u, \hat{\bm{x}}_{u \rightarrow u})
+ (1 - \lambda) \cdot \lVert \bm{x}_u - \hat{\bm{x}}_{u \rightarrow u} \rVert_1 
\big]
\end{align*}

To enforce the preservation of anatomical integrity and luminance during translation, we also introduce a Structural Consistency Loss $L_{\text{struct}}^{x}$ and a Luminance Consistency Loss $L_{\text{lum}}^{x}$ as:

\begin{align*}
L_{\text{struct}}^{x} &= \mathbb{E}_{\bm{x}_u, \bm{x}_l} \big[ 
1 - s(\hat{\bm{x}}_{u \rightarrow u}, \hat{\bm{x}}_{u \rightarrow l})
\big] \\
L_{\text{lum}}^{x} &= \mathbb{E}_{\bm{x}_u, \bm{x}_l} \big[ 
1 - l(\bm{x}_{l}, \hat{\bm{x}}_{u \rightarrow l}) 
\big]
\end{align*}

Additional loss components are adapted from the AGUIT framework~\cite{li2019attribute}, including: Brain Structure Adversarial Loss, Style Continuity Loss, Content-Style Separation Loss, Image Adversarial Loss, Image Classification Loss, and Feature Consistency Loss.
The overall objective for \( E_s, E_b, G \) and for \( D_{a}^{b}, D_{a}^{x}, D_{p}^{b}, D_{p}^{x} \) become respectively:

\begin{align*}
\text{L}_{G,E_b,E_s} &= \lambda_{\text{lum}}^{x} \text{L}_{\text{lum}}^{x} + \lambda_{\text{struct}}^{x} \text{L}_{\text{struct}}^{x} + \lambda_{\text{cla}}^{s} \text{L}_{\text{cla}}^{s} + \lambda_{\text{adv}}^{b} \text{L}_{\text{adv}}^{b} - \lambda_{\text{pre}}^{b} \text{L}_{\text{pre}}^{b} + \lambda_{\text{rec}}^{x} \text{L}_{\text{rec}}^{x} \\
&\quad + \lambda_{\text{adv}}^{x} \text{L}_{\text{adv}}^{x} + \lambda_{\text{pre}}^{x,G} \text{L}_{\text{pre}}^{x,G} + \lambda_{\text{cc}}^{x} \text{L}_{\text{cc}}^{x} + \lambda_{\text{lat}} \text{L}_{\text{lat}} \\
\text{L}_D &= - \lambda_{\text{adv}}^{c} \text{L}_{\text{adv}}^{b} + \lambda_{\text{pre}}^{b} \text{L}_{\text{pre}}^{b} - \lambda_{\text{adv}}^{x} \text{L}_{\text{adv}}^{x} + \lambda_{\text{pre}}^{x,D} \text{L}_{\text{pre}}^{x,D}
\end{align*}

\subsection{\bf Inference}
\label{sec:Inference}
\begin{figure}[ht]
\centering
\begin{overpic}[width=0.95\linewidth]{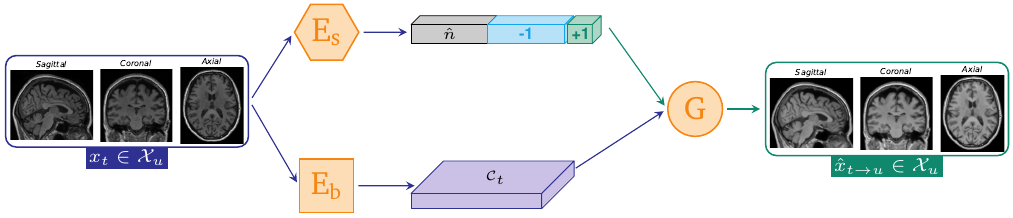}
    \put(0,1){\normalsize (a)}
\end{overpic}

\vspace{1.2em} 

\begin{overpic}[width=0.95\linewidth]{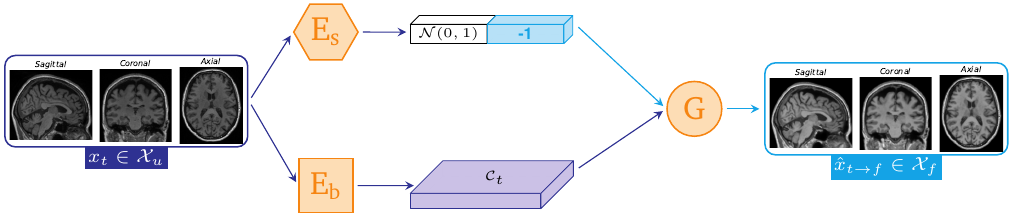}
    \put(0, 1){\normalsize (b)}
\end{overpic}

\caption{Visualizations of image generation during inference. (a) Generation with target scanner-site attributes. (b) Generation using style-agnostic.}
\label{fig:inf}
\end{figure}

At inference, a new image \( x_t \in \mathcal{X} \) is encoded by \( E_s \) and \( E_b \) to obtain its style \( s_t \) and anatomical structure \( c_t \). The attribute component of \( s_t \) can be modified—for instance, by setting values corresponding to a target scanner or site to \( +1 \)—to synthesize an image that retains the anatomy of \( x_t \) while exhibiting the desired scanner-site characteristics (Figure~\ref{fig:inf}a). Alternatively, the noise component can be sampled from \( z \sim \mathcal{N}(0, 1) \), with all attributes set to \( -1 \), enabling the generation of images with consistent scanner-site effects independent of any training data context (Figure~\ref{fig:inf}b).

\section{Data Acquisition and Preprocessing}
\label{sec:DATA}

\begin{table}[h]
\footnotesize
\centering
\caption{Overview of the training and test datasets used in this study. The table lists scanner models, manufacturers, the number of acquisition sites, dataset sources, and image counts. Training and test sets for healthy controls (\textit{HC}) are shown, along with the test sets containing both ASD, AD, and HC participants. The "Images (n)" column indicates the number of images used.}
\begin{tblr}{
  colspec = {@{} >{\raggedright\arraybackslash}p{3.5cm} >{\raggedright\arraybackslash}p{3.5cm} >{\raggedright\arraybackslash}p{2.1cm} >{\centering\arraybackslash}p{1.6cm} >{\raggedright\arraybackslash}p{2.0cm} >{\centering\arraybackslash}p{1.6cm} @{}},
  row{3,5,8,10,12,14,16,18,21,23,25,27,29,31} = {LightBlue}, 
}
\hline[0.5pt]
\textbf{} & \textbf{Scanner Model} & \textbf{Manufacturer} & \textbf{Sites (n)} & \textbf{Dataset} & \textbf{Images (n)} \\
\hline \hline
 \SetCell[r=5]{c,3.0cm}{\centering \textbf{Training} \\ \hspace{1.1cm} (\textit{HC})} & Prisma Fit & SIEMENS & 3 & ADNI 3 & 69 \\
\SetRow{bg=LightBlue}  & Prisma & SIEMENS & 5 & ADNI 3 & 86 \\
& TrioTim & SIEMENS & 2 & SALD, PPMI & 190 \\
& Gyroscan Intera & Philips & 1 & IXI & 133 \\
& Intera & Philips & 1 & IXI & 177 \\
\hline
\SetCell[r=6]{c,3.0cm}{\centering \textbf{Test} \\ \hspace{1.1cm} (\textit{HC})} & Unknown & GE & 1 & IXI & 74 \\
& Achieva dStream & Philips & 3 & ADNI 3, PPMI & 31 \\
& Achieva & Philips & 5 & ADNI 3, PPMI & 27 \\
& Skyra & SIEMENS & 4 & ADNI3 & 38 \\
& Gyroscan Intera & Philips & 1 & IXI & 189 \\
& TrioTim & SIEMENS & 2 & SALD, PPMI & 345 \\
\hline
\SetCell[r=7]{c,2.7cm}{\centering \textbf{Test} \\ \hspace{0.5cm} (\textit{ASD + HC})} & Signa   & GE      & 2 & ABIDE I & 185 \\
& MR750   & GE      & 1 & ABIDE I & 36  \\
& Verio   & SIEMENS & 2 & ABIDE I & 84  \\
& Allegra & SIEMENS & 3 & ABIDE I & 277 \\
& Achieva & Philips & 2 & ABIDE I & 104 \\
& Intera  & Philips & 2 & ABIDE I & 94  \\
& TrioTim & SIEMENS & 5 & ABIDE I & 332 \\
\hline
\SetCell[r=12]{c,2.7cm}{\centering \textbf{Test} \\ \hspace{0.5cm} (\textit{AD + HC})} & Prisma & SIEMENS & 6 & ADNI 3-4 & 39 \\
& Prisma Fit & SIEMENS & 9 & ADNI 3-4 & 59 \\  
& Biograph mMR  & SIEMENS & 1 & ADNI 3 & 3  \\
& Skyra & SIEMENS & 6 & ADNI 3-4 & 54 \\
& Achieva & Philips & 3 & ADNI 3-4 & 16 \\
& Achieva dStream  & Philips & 2 & ADNI 3 & 3  \\
& Ingenia & Philips & 4 & ADNI 3-4 & 17 \\
& MR 7700 & Philips & 2 & ADNI 4 & 22 \\
& DISCOVERY MR750 & GE & 6 & ADNI 3-4 & 30 \\
& Signa Premier & GE & 2 & ADNI 3 & 9 \\
& Verio   & SIEMENS & 1 & ADNI 3 & 1  \\
& Vida  & SIEMENS & 1 & ADNI 4 & 9  \\
\hline[0.5pt]
\end{tblr}
\label{tab:train_test_datasets}
\end{table}

We conducted experiments using T1-weighted MRI data from five large-scale, publicly available datasets, comprising both healthy controls and diverse clinical populations across multiple acquisition sites and scanner platforms. Table~\ref{tab:train_test_datasets} summarizes all datasets, including scanner models, manufacturers, acquisition sites, sources, and image counts.

\paragraph{Healthy Control Cohorts.} We included data from four datasets containing healthy participants: the Alzheimer's Disease Neuroimaging Initiative (ADNI3) \cite{jack2008alzheimer}, the Parkinson's Progression Markers Initiative (PPMI) \cite{marek2011parkinson}, the IXI Brain Development Dataset (IXI) \cite{ixidata}, and the Southwest University Adult Lifespan Dataset (SALD) \cite{wei2018structural}. Together, these datasets include MRI scans acquired from 9 different scanners across 28 sites.

\paragraph{Clinical Cohorts.} We additionally incorporated T1-weighted MR images from two large multi-site datasets containing both clinical and healthy participants, providing a diverse and challenging evaluation setting. The first dataset, the Autism Brain Imaging Data Exchange I (ABIDE I) \cite{di2014autism}, includes structural MRI data from 1,112 individuals—comprising 539 participants with autism spectrum disorder (ASD) and 573 healthy controls. ABIDE I spans 17 international sites with heterogeneous acquisition protocols, scanner manufacturers, and demographic distributions, thus offering a rigorous benchmark for evaluating the robustness and generalizability of harmonization methods across diverse scanning environments.
The second dataset, the Alzheimer’s Disease Neuroimaging Initiative (ADNI; Phases 3–4) \cite{jack2008alzheimer}, includes participants diagnosed with Alzheimer’s disease (AD) as well as healthy controls. From these phases, we included a subset of 94 participants with AD and 168 healthy controls. ADNI 3–4 provides structural MRI data from 43 imaging sites, acquired using Siemens, Philips, and GE scanners, serving as a complementary benchmark for evaluating harmonization performance in neurodegenerative populations.

Together, these datasets enable a comprehensive evaluation of the proposed model across a wide range of acquisition conditions, scanner types, and population characteristics, effectively encompassing most real-world scanning environments encountered in both clinical and research settings. 

\paragraph{Image Preprocessing Pipeline.}
Image preprocessing was conducted using the FSL library \cite{fsl1, fsl2}. Initially, the images were standardized to a common orientation. Next, bias-field correction was applied to address magnetic field variations in the MRI scanner. The images were then registered to the Standard MNI152-T1-1mm space. The final preprocessed images had a shape of $(1, \ 182, \ 218, \ 182)$.

\section{Experimental setup}
\label{sec:EXP}
The model was trained on a dataset of 655 T1-weighted MR images from healthy controls, acquired across 5 scanners and 12 sites. For evaluation, we used an independent test set comprising 704 healthy control images collected using 6 scanners across 16 sites. Additional test sets included both clinical and healthy participants from the ABIDE I and ADNI 3–4 datasets. A detailed summary of the training and test datasets is presented in Table~\ref{tab:train_test_datasets}.

To evaluate our model, we harmonized all test images to match the domains of two distinct scanner-site pairs—referred to as A and B (Gyroscan Intera -- Guy’s Hospital, and TrioTim -- Southwest University, China)—as well as to a style-agnostic reference. The effectiveness of the harmonization was evaluated using three complementary criteria: (i) preservation of anatomical structures, (ii) consistency of image appearance across sites, and (iii) performance on downstream predictive tasks. In addition, axial, coronal, and sagittal slices were visually inspected to qualitatively assess the realism and inter-site consistency of the harmonized images.

\paragraph{Preservation of Anatomical Structure.}  We used the structural component of the SSIM to assess the preservation of anatomical structures following harmonization. Given the absence of ground truth data for contrast and luminance, we focus specifically on SSIM between the original and harmonized images. This metric quantifies the extent to which the underlying anatomical features are maintained after harmonization. Specifically, we computed SSIM values across image pairs representing diverse anatomical structures, as well as across the entire test set, for all three harmonization targets (scanner-site pairs A and B, and the style-agnostic reference), considering each harmonized–original image pair.

\paragraph{Consistency of image appearance.} We assess appearance consistency by comparing mean voxel intensity distributions across scanner-site pairs before and after harmonization. To quantify alignment, we compute the Wasserstein distance between all pairs of scanner-site distributions, both pre- and post-harmonization, for all three harmonization targets. We report the mean and standard deviation in both cases. Additionally, to further assess consistency, we calculate the SSIM, focusing on the luminance component, between all pairs of original test images and, separately, all pairs of harmonized images. Again, we report the mean and standard deviation before and after harmonization. These analyses were conducted using the test dataset composed exclusively of healthy controls (Table~\ref{tab:train_test_datasets}).

\paragraph{Downstream Analysis.} To evaluate the impact of harmonization on downstream neuroimaging applications, we performed two predictive tasks using convolutional neural networks (CNNs): (i) \textit{age prediction} and (ii) \textit{disease classification}. In both tasks, we employed 3D convolutional architectures based on the ResNet family. Each model was trained separately on non-harmonized and harmonized data to assess the effect of harmonization on predictive performance. Model evaluation was conducted using 5-fold cross-validation, reporting the mean and standard deviation of the respective metrics.

For \textit{age prediction}, the CNN was trained to estimate chronological age from structural MRI scans using the ABIDE I dataset. Performance was assessed via mean absolute error (MAE), root mean squared error (RMSE), and the coefficient of determination ($R^2$).

For \textit{disease classification}, a binary 3D ResNet model was trained to distinguish Alzheimer's disease (AD) patients from healthy controls using the ADNI 3–4 dataset. Evaluation metrics included accuracy, balanced accuracy, and area under the receiver operating characteristic curve (AUC).

\section{Results}
\label{sec:RESULTS}
In this section, we present the results of the experiments described in Section \ref{sec:EXP}. 

\begin{figure}[t!]
    \centering
    \includegraphics[width=\linewidth]{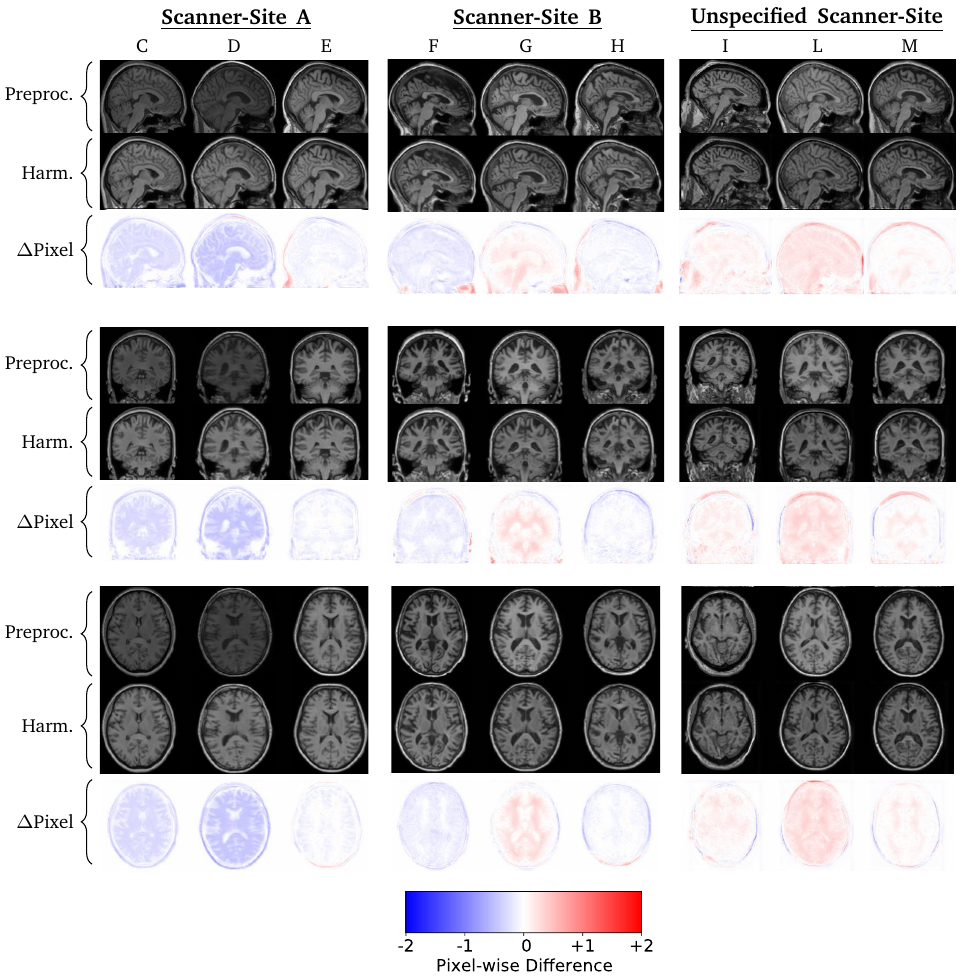}
    \caption{Axial, coronal, and sagittal slices from 9 MR images, both from the original scans (C-M) and their harmonized versions according to scanner-site pair A (left column), scanner-site pair B (middle column), and a style-agnostic target (right column). Heatmaps indicate pixel-wise differences between the original and harmonized images.}
    \label{fig:montage}
\end{figure}

Figure~\ref{fig:montage} provides a visual comparison of the harmonization outcomes across the three target domains. The original images exhibit substantial inter-domain variability, whereas the harmonized outputs demonstrate visually consistent appearances across all style targets. This consistency is particularly evident in the heatmaps, which illustrate pixel-wise differences. The anatomical structures remain visually well-preserved after harmonization, indicating that the proposed method effectively reduces style-related variability while maintaining underlying anatomical integrity.

Nevertheless, visual assessment alone is insufficient to draw definitive conclusions regarding the quality of harmonization or the preservation of anatomical details. Therefore, we complement the qualitative evaluation with quantitative analyses to rigorously assess both the harmonization performance and the fidelity of anatomical preservation.

\paragraph{Preservation of Anatomical Structure.} Across 100 randomly selected image pairs representing diverse anatomical structures, the average SSIM was $0.68 \pm 0.05$. When all test images and all three harmonization targets were considered, the average SSIM increased to $0.971 \pm 0.012$. This marked improvement demonstrates that the proposed model achieves strong anatomical preservation, reflected in the significantly higher SSIM scores between original and harmonized images of the same subject.

\begin{figure}
    \centering
    \includegraphics[width=\textwidth]{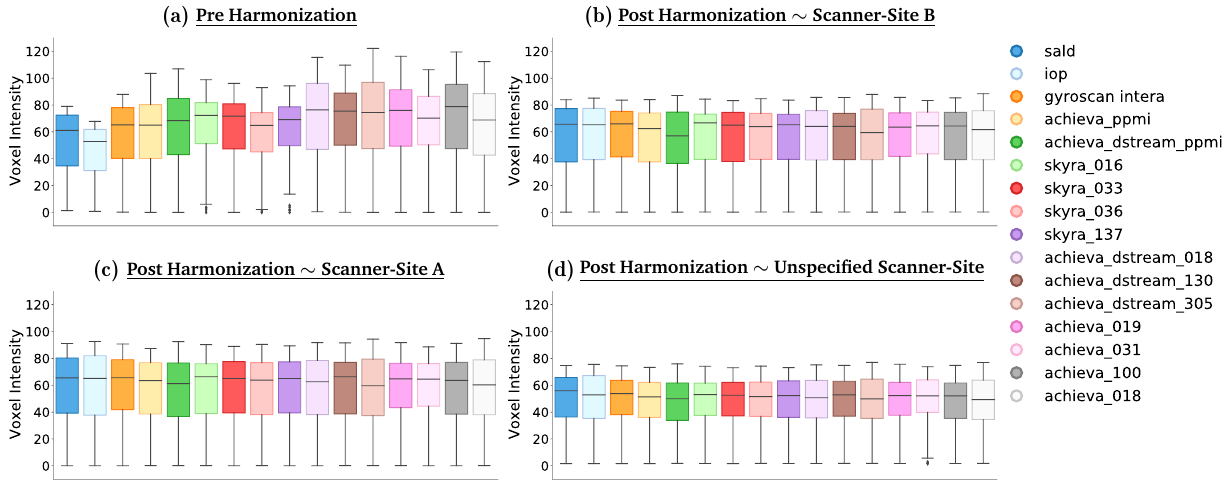}
    \caption{Comparison of voxel intensity mean distributions for each scanner-site pair in the test images, before (a) and after harmonization using the scanner-site pair B (b), the scanner-site pair A (c), and the unspecified scanner-site as the target scanner-site attributes.}
    \label{fig:distr}
\end{figure}

\paragraph{Consistency of image appearance.} We compared the voxel intensity mean distributions for each of the 16 scanner-site pairs in the test images, both before and after harmonization with respect to the three target attributes (Figure \ref{fig:distr}). As shown in the plots, the proposed model achieves strong alignment of voxel intensity distributions across all target types, consistent with visual inspection.

\begin{figure}[h]
    \centering
    \includegraphics[width=\linewidth]{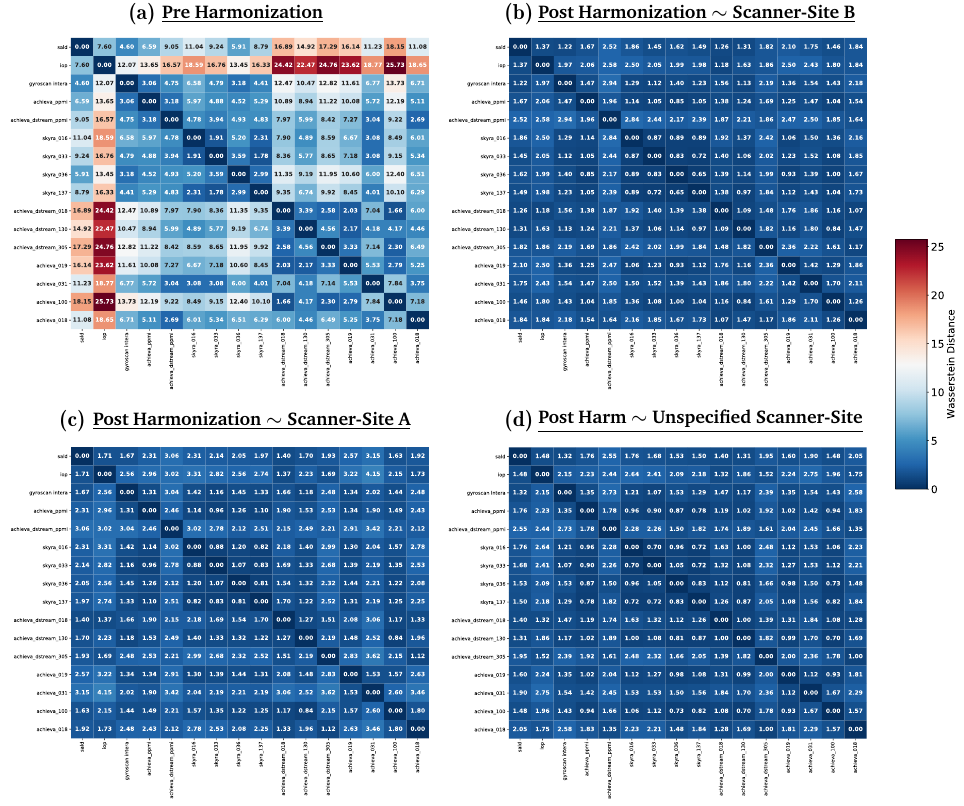}
    \caption{Heatmaps showing the pairwise Wasserstein distances between mean voxel intensity distributions for all scanner-site pairs in the healthy control test dataset. Panel (a) shows distances before harmonization, while panels (b), (c), and (d) show distances after harmonization to scanner-site pair B, scanner-site pair A, and the style-agnostic target, respectively.}
    \label{fig:wass_heatmaps}
\end{figure}

This observation is quantitatively supported by the results summarized in Table~\ref{tab:table_wass_lum}. The Wasserstein distance between scanner-site distributions is markedly reduced after harmonization, from $8.454 \pm 5.347$ pre-harmonization to $1.989 \pm 0.703$, $1.601 \pm 0.492$, and $1.677 \pm 0.559$ for targets A, B, and the style-agnostic reference, respectively. This substantial reduction indicates that intensity distributions across scanners and sites are brought into close agreement, effectively mitigating scanner- and site-related variability. Corresponding heatmaps in Figure~\ref{fig:wass_heatmaps} visualize the pairwise Wasserstein distances.

Similarly, the luminance similarity increases from $0.952 \pm 0.037$ pre-harmonization to $0.981 \pm 0.018$, $0.986 \pm 0.016$, and $0.989 \pm 0.014$ after harmonization to the three targets. These results demonstrate that the harmonized images not only exhibit well-aligned voxel intensity distributions but also maintain high voxel-level luminance consistency across sites.

\begin{table}[h]
\footnotesize
\renewcommand{\arraystretch}{1.5}
\centering
\caption{Evaluation of luminance similarity between images and Wasserstein distance between the mean voxel intensity distribution. We report the values before and after harmonization to all three targets considered.}
\begin{tblr}{@{} 
>{\raggedright\arraybackslash}m{2.9cm} 
>{\centering\arraybackslash}m{2.9cm}   
>{\centering\arraybackslash}m{2.7cm}   
>{\centering\arraybackslash}m{2.7cm}   
>{\centering\arraybackslash}m{2.7cm}   
@{}}
\cline{1-5}
\SetCell{c}{\textbf{Metric}} & 
\SetCell{c}{\textbf{Pre-Harmonization}} & 
\SetCell{c}{\textbf{A} \hspace{1cm} \\ (\textit{Scanner-Site})} &  
\SetCell{c}{\textbf{B} \hspace{1cm} \\ (\textit{Scanner-Site})} &  
\SetCell{c}{\textbf{Unspecified} \\ (\textit{Scanner-Site})}  \\
\hline
\textit{Wasserstein distance} & $8.454 \pm 5.347$ & $1.989 \pm 0.703$ & $1.601 \pm 0.492$ & $1.677 \pm 0.559$     \\
\textit{Luminance Similarity} & $0.952 \pm 0.037$ & $0.981 \pm 0.018$ & $0.986 \pm 0.016$ & $0.989 \pm 0.014$   \\
\hline [0.5pt]
\end{tblr}
\label{tab:table_wass_lum}
\end{table}

\paragraph{Downstream Analysis.} For \textit{age prediction}, Table~\ref{tab:table_age_pred} shows that the mean absolute error (MAE) decreases from $5.36 \pm 1.89$ years pre-harmonization to $3.30 \pm 0.48$, $3.39 \pm 0.36$, and $3.31 \pm 0.78$ years for the three harmonization targets (scanner-site A, scanner-site B, and unspecified target, respectively). Similarly, the coefficient of determination ($R^2$) increases from $0.091 \pm 0.571$ to $0.643 \pm 0.051$, $0.545 \pm 0.094$, and $0.639 \pm 0.140$, while the root mean squared error (RMSE) similarly decreases across all targets. These improvements indicate that harmonization effectively minimizes scanner- and site-related variability, allowing the CNN to extract biologically relevant age-related information without interference from acquisition-related biases.

For \textit{Alzheimer’s disease (AD) classification}, harmonization leads to clear improvements (Table~\ref{tab:table_ad_pred}). The area under the ROC curve (AUC) increases from $0.778 \pm 0.062$ pre-harmonization to $0.848 \pm 0.082$, $0.822 \pm 0.077$, and $0.836 \pm 0.072$ for scanner-site A, scanner-site B, and the unspecified target, respectively. Other metrics, including accuracy and balanced accuracy, show similar improvements, reflecting enhanced overall model performance across all harmonization targets.

The consistency of these improvements across both tasks and all harmonization targets demonstrates the robustness of the approach. Overall, these downstream analyses indicate that harmonization not only improves inter-site consistency and preserves structural information but also enhances the predictive utility for both regression and classification. This confirms that the method effectively reduces site-specific biases while retaining biologically relevant information critical for neuroimaging applications.

\begin{table}[h]
\footnotesize
\renewcommand{\arraystretch}{1.5}
\centering
\caption{Performance of the age prediction model before and after data harmonization. The table reports mean absolute error (MAE), root mean squared error (RMSE), and coefficient of determination ($R^2$) with standard deviations across folds.}
\begin{tblr}{@{} 
>{\raggedright\arraybackslash}m{1.0cm} 
>{\centering\arraybackslash}m{2.9cm}   
>{\centering\arraybackslash}m{2.4cm}   
>{\centering\arraybackslash}m{2.4cm}   
>{\centering\arraybackslash}m{2.4cm}   
@{}}
\cline{1-5}
\SetCell{c}{\textbf{Metric}} & 
\SetCell{c}{\textbf{Pre-Harmonization}} & 
\SetCell{c}{\textbf{A} \hspace{1cm} \\ (\textit{Scanner-Site})} &  
\SetCell{c}{\textbf{B} \hspace{1cm} \\ (\textit{Scanner-Site})} &  
\SetCell{c}{\textbf{Unspecified} \\ (\textit{Scanner-Site})}  \\
\hline
\textit{MAE} & $5.36 \pm 1.89$ & $3.30 \pm 0.48$ & $3.39 \pm 0.36$ & $3.31 \pm 0.78$   \\
\textit{RMSE} & $7.20 \pm 2.38$ & $4.79 \pm 0.46$ & $5.41 \pm 0.80$ & $4.74 \pm 1.05$     \\
\textit{R}$^2$ & $0.091 \pm 0.571$ & $0.643 \pm 0.051$ & $0.545 \pm 0.094$ & $0.639 \pm 0.140$     \\
\hline [0.5pt]
\end{tblr}
\label{tab:table_age_pred}
\end{table}

\begin{table}[h]
\footnotesize
\renewcommand{\arraystretch}{1.5}
\centering
\caption{Classification performance (AD vs Healthy Controls) on MRI data before and after harmonization. The table reports the mean and standard deviation across folds for accuracy, balanced accuracy, and AUC.}
\begin{tblr}{@{} 
>{\raggedright\arraybackslash}m{2.4cm} 
>{\centering\arraybackslash}m{2.9cm}   
>{\centering\arraybackslash}m{2.4cm}   
>{\centering\arraybackslash}m{2.4cm}   
>{\centering\arraybackslash}m{2.4cm}   
@{}}
\cline{1-5}
\SetCell{c}{\textbf{Metric}} & 
\SetCell{c}{\textbf{Pre-Harmonization}} & 
\SetCell{c}{\textbf{A} \hspace{1cm} \\ (\textit{Scanner-Site})} &  
\SetCell{c}{\textbf{B} \hspace{1cm} \\ (\textit{Scanner-Site})} &  
\SetCell{c}{\textbf{Unspecified} \\ (\textit{Scanner-Site})}  \\
\hline
\textit{Accuracy} & $0.702 \pm 0.062$ & $0.759 \pm 0.095$ & $0.736 \pm 0.052$ & $0.744 \pm 0.065$ \\
\textit{Balanced Accuracy} & $0.647 \pm 0.062$ & $0.727 \pm 0.091$ & $0.667 \pm 0.067$ & $0.704 \pm 0.059$ \\
\textit{AUC} & $0.778 \pm 0.062$ & $0.848 \pm 0.082$ & $0.822 \pm 0.077$ & $0.836 \pm 0.072$ \\
\hline [0.5pt]
\end{tblr}
\label{tab:table_ad_pred}
\end{table}

\section{Conclusion}
\label{sec:CONCLUSION}
We have presented a novel 3D T1-weighted MRI harmonization framework that disentangles anatomical content from scanner- and site-specific variations. By incorporating an SSIM-based loss alongside traditional reconstruction and adversarial objectives, our method achieves high structural fidelity while effectively reducing inter-site variability. Extensive evaluation across multiple large-scale datasets—including both healthy controls and clinical populations (ASD and AD)—demonstrated that harmonized images exhibit consistent voxel intensity distributions, improved luminance similarity, and preserved anatomical structures.

Importantly, harmonization led to substantial improvements in downstream neuroimaging tasks: brain age prediction errors decreased markedly, and Alzheimer’s disease classification performance improved across all harmonization targets. These results confirm that our framework not only standardizes MRI appearance across scanners and sites but also enhances the extraction of biologically meaningful features relevant for clinical and research applications.

Overall, our approach provides a robust, generalizable solution for large-scale, multicenter neuroimaging studies, facilitating reproducible analyses and cross-site comparability. Future work will explore the extension of this framework to other imaging modalities and anatomical regions, as well as the integration of additional covariates—such as age, sex, and pathological markers—into the style representation.

\clearpage
\phantomsection

\section*{Acknowledgements}
L. Caldera is funded by Health Big Data project sponsored by the Italian Ministry of Health (CCR-2018-23669122). L. Cavinato is funded by the National Plan for NRRP Complementary Investments “Advanced Technologies for Human-centred Medicine” (PNC0000003). The present research is part of the activities of “Dipartimento di Eccellenza 2023-2027".

Data collection and sharing for this project was funded by the Alzheimer's Disease Neuroimaging Initiative (ADNI) (National Institutes of Health Grant U01 AG024904) and DOD ADNI (Department of Defense award number W81XWH-12-2-0012). ADNI is funded by the National Institute on Aging, the National Institute of Biomedical Imaging and Bioengineering, and through generous contributions from the following: AbbVie, Alzheimer’s
Association; Alzheimer’s Drug Discovery Foundation; Araclon Biotech; BioClinica, Inc.; Biogen; Bristol-Myers Squibb Company; CereSpir, Inc.; Cogstate; Eisai Inc.; Elan Pharmaceuticals, Inc.; Eli Lilly and Company; EuroImmun; F. Hoffmann-La Roche Ltd and its affiliated company Genentech, Inc.; Fujirebio; GE Healthcare; IXICO Ltd.; Janssen Alzheimer Immunotherapy Research \& Development, LLC.; Johnson \& Johnson
Pharmaceutical Research \& Development LLC.; Lumosity; Lundbeck; Merck \& Co., Inc.; Meso Scale Diagnostics, LLC.; NeuroRx Research; Neurotrack Technologies; Novartis
Pharmaceuticals Corporation; Pfizer Inc.; Piramal Imaging; Servier; Takeda Pharmaceutical Company; and Transition Therapeutics. The Canadian Institutes of Health Research is
providing funds to support ADNI clinical sites in Canada. Private sector contributions are facilitated by the Foundation for the National Institutes of Health (www.fnih.org). The grantee organization is the Northern California Institute for Research and Education, and the study is coordinated by the Alzheimer’s Therapeutic Research Institute at the University of Southern California. ADNI data are disseminated by the Laboratory for Neuro Imaging at the University of Southern California.

\bibliographystyle{unsrt}  


\end{document}